\documentclass[12pt,a4paper,oneside]{article}

\usepackage[top=3cm, bottom=3cm, left=2cm, right=2cm]{geometry}

\usepackage{graphicx}
\usepackage{verbatim}
\usepackage{amsmath}
\usepackage{eepic}
\usepackage{bbold}

\begin{document}

\title{The computation of first order moments \\ on junction trees}
\date{}

\author{Milo\v s B. Djuri\'c,
        Velimir M. Ili\'c,
        and Miomir S. Stankovi\'c
}

\maketitle

\newtheorem{definition}{\textbf{Definition}}
\newtheorem{lemma}{\textbf{Lemma}}
\newtheorem{example}{\textbf{Example}}

\begin{abstract}
We review some existing methods for the computation of first order
moments on junction trees using Shafer-Shenoy algorithm. First, we
consider the problem of first order moments computation as
vertices problem in junction trees. In this way, the problem is
solved using the memory space of an order of the junction tree
edge-set cardinality. After that, we consider two algorithms,
Lauritzen-Nilsson algorithm, and Mau{\'a} et al. algorithm, which
computes the first order moments as the normalization problem in
junction tree, using the memory space of an order of the junction
tree leaf-set cardinality.

\end{abstract}

\newcommand{\bs}{\boldsymbol}
\newcommand{\mb}{\mathbf}
\newcommand{\mc}{\mathcal}
\newcommand{\bc}[2]{\binom{\ #1 \ }{\ #2 \ }}

\newcommand{\sV}{\mc V}
\newcommand{\sE}{\mc E}
\newcommand{\sM}{\mc M}

\newcommand{\tr}{\text r}
\newcommand{\tq}{\text q}
\newcommand{\tbc}{\text{bc}}
\newcommand{\tZ}{\text Z}
\newcommand{\tsr}{\text{sr} }

\newcommand{\sR}{\mathbb R}
\newcommand{\sN}{\mathbb N}

\newcommand{\sK}{\mc K}
\newcommand{\sA}{\mc A}
\newcommand{\sX}{\mc X}
\newcommand{\sT}{\mc T}
\newcommand{\sC}{\mc V}
\newcommand{\sS}{\mc S}

\newcommand{\vg}{\bs g}
\newcommand{\vx}{\bs x}
\newcommand{\vt}{\bs t}

\newcommand{\ra}{\rightarrow}
\newcommand{\da}{\downarrow}
\newcommand{\sm}{\setminus}
\newcommand{\es}{\emptyset}

\newcommand{\T}[2]{\text T^{#1}_{#2}}
\newcommand{\M}[2]{\text M^{#1}_{#2}}

\newcommand{\egf}[2]{\sum_{\bs #1 \in \Ainf} \frac{#2^{(#1)}}{#1 !} \cdot \vt^{\bs #1}}
\newcommand{\gf}[2]{\sum_{\bs #1 \in \Ainf} #2^{(#1)} \cdot \vt^{\bs #1}}
\newcommand{\bsel}[3]{\big(\ #1^{(\bs #2)}\ \big)_{#2 \in #3} }
\newcommand{\Tup}[2]{\Big(\ #1 \ \Big)_{#2}}
\newcommand{\tup}[2]{\big(\ #1 \ \big)_{#2}}

\newcommand{\mpa}{\text{mpa}}

\newcommand{\bsprod}[5]{\Big(\ \sum_{\bs #1 \leq \bs #2} \bc{\bs #2}{\bs #1}\ #3^{(\bs #1)} \cdot #4^{(\bs #2 - \bs #1)} \Big)_{\bs #2 \in #5}}
\newcommand{\bsprcor}[4]{\sum_{\bs #1 \leq \bs #2} \bc{\bs #2}{\bs #1}\ #3^{(\bs #1)} \cdot #4^{(\bs #2 - \bs #1)}}

\newcommand{\Anu}{A_{\bs \nu}}

\newcommand{\mia}{\bs \alpha}
\newcommand{\mib}{\bs \beta}
\newcommand{\mig}{\bs \gamma}
\newcommand{\minu}{\bs \nu}
\newcommand{\Ainf}{\sN_0^d}

\newcommand{\e}{\bs{e}}
\newcommand{\Dop}[2]{\mc D^{(#1)} \big\{\ #2 \ \big\}_{\vt =
0}}
\newcommand{\mgf}[2]{M_{#1,#2}(\vt)}
\newcommand{\mom}[3]{\mu_{#1, #2}^{(#3)}}

\newcommand{\mgffac}[2]{\ #1 \cdot \e^{#2 \cdot\vt}}

\newcommand{\hbegf}[3]{\mc B^{(#1)} \big\{\ \egf{#2}{#3} \ \big\}}
\newcommand{\hpegf}[3]{\mc P^{(#1)} \big\{\ \gf{#2}{#3} \ \big\}}

\newcommand{\hp}{\mc P}

\newcommand{\hB}[2]{\mc B^{(#1)} \Big\{\ #2 \ \Big\}}
\newcommand{\hb}[2]{\mc B^{(#1)} \big\{\ #2 \ \big\}}

\newcommand{\hfP}[2]{\mc P^{(#1)} \Big\{\ #2 \ \Big\}}
\newcommand{\hfp}[2]{\mc P^{(#1)} \big\{\ #2 \ \big\}}

\newcommand{\hbgf}[3]{\mc B^{(#1)} \big\{\ \gf{#2}{#3} \ \big\}}

\renewcommand{\dim}{d}
\newcommand{\pd}{\partial}


\section{Shafer-Shenoy algorithm}


\subsection{Potentials and operations}

Let $X = (X_u : u \in U)$ be a finite collection of discrete
random variables. Let $\Omega_u$ denotes the set of possible
values that $X_u$ can take. For $A \subseteq U$ we write
$\Omega_A$ for the Cartesian product $\times_{u \in A} \Omega_u$
and write $X_A$ for $\{X_u : u \in A\}$ \footnote{We implicitly
assume some natural ordering in sets}. The elements of $\Omega_A$,
$A \subseteq U$ are denoted $x_A$ and called the
\emph{configuration}. We adopt the convention that $\Omega_\es$
consists of a single element $x_\es = \diamond$, i.e. $\Omega_\es
=\{ \diamond \}$.

Let $\Pi_U$ be a set of functions on $\pi_A: \Omega_A \ra \sK$,
where $\ A \subseteq U$, i.e. $\Pi_U = \big\{ \pi_A: \Omega_A \ra
\sK\ |\ A \subseteq U \big\}$. In the following text, the
functions from $\Pi_U$ are called potentials. Let $\otimes$ be a
binary operation on $\sK$ called combination and let $^{\da}$
denotes the external operation called marginalization which to
every $\pi_A: \Omega_A \ra \sK$ associates $\pi_{A}^{\da B}:
\Omega_{A\cap B} \ra \sK$, where $A,B \in U$.

We assume that the following Shafer-Shenoy axioms hold for
combination and marginalization.

\subsection{Shafer-Shenoy axioms}

\textbf{Axiom 1 (Commutativity and Associativity)} Let $\pi_{A},
\pi_{B}$ and $\pi_{C}$ be potentials. Then

\begin{equation}
\pi_{A}\otimes \pi_{B} = \pi_{B}\otimes \pi_{A} \quad \text{and}
\quad \pi_{A}\otimes (\pi_{B}\otimes \pi_{C}) = (\pi_{A}\otimes
\pi_{B})\otimes \pi_{C}.
\end{equation}

Axiom 1 allows us to use the notation $\pi_{A}\otimes
\pi_{B}\otimes \pi_{C}$.

\noindent \textbf{Axiom 2 (Consonance)} Let $\pi_A$ be a potential
on A, and let $A\supseteq B\supseteq C$. Then

\begin{equation}
\label{problem: consonance} (\pi_A^{\downarrow B})^{\downarrow C}
= \pi_A^{\downarrow C}.
\end{equation}

\noindent \textbf{Axiom 3 (Distributivity)} Let $\pi_{A}$ and
$\pi_{B}$ be potentials on $A$ and $B$, respectively. Then

\begin{equation}
(\pi_{A}\otimes \pi_{B})^{\downarrow A} = \pi_{A}\otimes
\pi_{B}^{\downarrow A}.
\end{equation}

\subsection{Junction Tree}

The joint potential $\pi_U: \Omega_U \ra \sK$ is said factorize on
$\sT$ with respect to $\otimes$ if there exists potentials $\pi_V:
\Omega_V \ra \sK$ for $V \in \sV$, so that we can write $\pi$ as
\begin{equation}
\label{problem: piU=prod piC}
\pi_U = 
\bigotimes_{V \in \mc V}\pi_V.
\end{equation}
In this paper we consider the joint potentials which can be
represented with \emph{junction tree} which is defined as follows.

\begin{definition}
Let $\sV = \big\{ V_1, V_2, \dots, V_n \big\}$ be a
collection of subsets of $U$ 
(set of variable indices) and $\mathcal T$ a tree with $\sV$ as
its node set (corresponds to a set of local domains). Then
$\mathcal T$ is said to be a \emph{junction tree} if any
intersection $V_i \cap V_j$ of a pair $V_i, V_j$ of sets in $\sV$
is contained in every node on the unique path in $\mathcal T$
between $V_i$ and $V_j$. Equivalently, for any $u \in U$, the set
of subsets in $\sV$ containing $u$ induces a connected subtree of
$\mathcal T$.
\end{definition}

The set of all neighbors of $A$ is denoted $ne(A)$. We omit the
parentheses from the notation when it is not prone to
misunderstanding. Hence, $ne(A)\sm B$ stands for the set of all
neighbors of $A$ without $B$, $ne(A)\sm B, C$ for the set of all
neighbors of $A$ without $B$ and $C$ and so on. $V_i \sim V_{i+1}$
denotes that $V_i$ and $V_{i+1}$ are neighbors.

A junction tree is usually drawn with sets $V_i$ as node labels.
In the following text the node will be identified with the label.
The general procedure for the junction tree building can be found
in \cite{Aji_McEliece_00} and \cite{Cowell_et_al_99}.  An example
of the junction tree which corresponds to \emph{chain
factorization},
\begin{equation}
\pi_U = \bigotimes_{i=1}^n \pi_{V_i},\quad%
V_i \sim V_{i+1} \ \text{for} \ i= 1, \dots, n-1,
\end{equation}
is given in Fig. 1.

\begin{figure}[h]
\label{JT_chain} \centering \setlength{\unitlength}{1mm}

\begin{picture}(78, 12)


  \put(5, 5){\circle{10}}
  \put(3, 4){$V_1$}


  \put(10,5){\line(1,0){10}}

  \put(25, 5){\circle{10}}
  \put(23, 4){$V_2$}


    \put(30,5){\line(1,0){5}}
    \put(37,4){$\cdots$}
    \put(43,5){\line(1,0){5}}

  \put(53, 5){\circle{10}}
  \put(49.5, 4){$V_{n-1}$}

  \put(58,5){\line(1,0){10}}

  \put(73, 5){\circle{10}}
  \put(71, 4){$V_n$}

\end{picture}

\center Fig. 1.\ The junction tree for chain factorization $\pi_U
=
\bigotimes_{i=1}^n \pi_{V_i}$, $V_{i} \sim V_{i+1}$.
\end{figure}

\subsection{Problems}

The junction tree enables the solution of three
important problems:

\begin{enumerate}

\item The \textit{single vertex problem} at node $A$ is defined as the computation of
the potential $\psi_A: \Omega_A \ra \mc K$, defined by
\begin{equation}
\label{JTA MPF}%
\psi_A = \pi_U^{\da A} = \Big( \bigotimes_{V \in \mc V}\pi_V(x_V)
\Big)^{\da A},
\end{equation}

\item The \textit{all vertices problem} is defined as the computation of the functions $\psi_A$ for all $A \in \sC$;

\item The \textit{normalization problem} is the marginalization of the joint potential (\ref{problem: piU=prod piC})
to the empty set $\emptyset$. Using the consonance of the
marginalization (\ref{problem: consonance}), it can straightforwardly be solved by the
solution of the single vertex problem in arbitrary node $A$:

\begin{equation}
\label{JTA totsum u(y)}%
\pi_U^{\da \emptyset} = \big(\pi_U^{\da A}\big)^{\da \emptyset}=
\psi_A^{\da \emptyset}
\end{equation}

\end{enumerate}

\subsection{Local computation algorithm}

These problems can efficiently be solved with the
\emph{Shafer-Shenoy local computation algorithm} (\emph{LCA}). The
algorithm can be described as passing the messages over the edges
and processing the messages in the nodes of the junction tree.

Messages are passed between the vertexes from $\sV$ via mailboxes.
All mailboxes are initialized as empty. When a message has been
placed in a mailbox, the box is full. A node $A$ in the junction
tree is allowed to send a message to its neighbor $B$ if it has
not done so before and if all $A$-incoming mailboxes are full
except possibly the one which is for $B$-outgoing messages. So,
initially only leaves (nodes which have only one neighbor) of the
junction tree are allowed to send messages. But as the message
passing proceeds, other nodes will have their turn and eventually
all mailboxes will be full, i.e., exactly two messages will have
been passed along each branch of the junction tree.

The message form $A$ to $B$ is a function $\pi_{A \ra B}:
\Omega_{A \cap B} \ra \sK $. The passage of a message $\pi_{A\ra
B}$ from node $A$ to node $B$ is performed by absorption.
Absorption from clique $A$ to clique $B$ involves eliminating the
variables $A \sm B$ from the potentials associated with $A$ and
its neighbors except $B$. The structure of the message $\pi_{A\ra
B}$ is given by
\begin{equation}
\label{Message_A_B}
\pi_{A\ra B} = \Big( \pi_A\otimes(\bigotimes_{C\in ne(A)\setminus
B} \pi_{C\ra A})\Big)^{\downarrow B}.
\end{equation}
where $\pi_{C\ra A}$ is the message passed from $C$ to $A$. Since
the leaves has only one neighbor, the product on the righthand
side is empty and the message can initially be computed as
\begin{equation}
\pi_{A\ra B}=\pi_A^{\downarrow B}.
\end{equation}

Suppose we start with a joint potential $\pi_U$ on a junction tree
$\sT$, and pass messages towards a root clique $R$ as described
above. When $R$ has received a message from each of its neighbors,
the combination of all messages with its own potential is equal to
a decomposition of the $R$-marginal of $\pi_U$.
\begin{equation}
\pi_U^{\downarrow R} =%
\Big( \bigotimes_{V \in \sC} \pi_V \Big)^{\da R} =  \pi_R
\otimes\Big(\bigotimes_{V\in ne(R)} \pi_{V \ra R}\Big).
\end{equation}
where $\sC$ is vertex-set in $\sT$.

Hence, if we want to solve the single vertex problem at node $A$,
we need to compute all messages incoming to $A$, while for the all
vertices problem we need the messages between all pairs of nodes
in the tree.

For the single vertex problem, the algorithm starts at the leaves
which send the messages to their neighbors. A node sends a message
to a neighbor, once it has received messages from each of its
other neighbors. The node $A$ never sends a message. Thus, each
message is sent only once until $A$ has received the messages from
all the neighbors at which point the required marginal is computed
and the algorithm terminates with the total number of computed
messages equal to the number of edges of the tree. Once we have
solved the single vertex problem in the node $A$, the
normalization problem can be solved with (\ref{JTA totsum u(y)}).

The first part of the algorithm for all vertices problem is
similar to the single vertex case. The messages are sent from
leaves toward the tree until a node $C$ has received the messages
from all the neighbors. After that the messages are sent from $C$
to the leaves. The algorithm stops when all leaves receive
messages. The total number of computed messages is equal two times
the number of edges in the tree (for any two nodes $A$ and $B$ we
send the messages $\pi_{A \ra B}$ and $\pi_{B \ra A}$).

\section{First order moments}

\subsection{Operations on the set of functions}

For real-valued functions $\phi_A: \Omega_A \ra \sR$ and $\phi_B:
\Omega_B \ra \sR$ the sum, $\phi_A + \phi_B: \Omega_{A \cup B} \ra
\sR$ and the product, $\phi_A \cdot \phi_B: \Omega_{A \cup B} \ra
\sR$, are respectively defined with:
\begin{align}
\big(\phi_A + \phi_B\big)(x_{A \cup B}) &= \phi_A(x_A)+\phi_B(x_B) \\
\big(\phi_A \cdot \phi_B\big)(x_{A \cup B}) &= \phi_A(x_A) \cdot \phi_B(x_B)
\end{align}
for all $x_A \in \Omega_A$ and $x_B \in \Omega_B$. The product
$\phi_A \cdot \phi_B$ is, usually, shortly denoted with
$\phi_A\phi_B$.

We define \emph{sum-marginal} operator $\sum_{x_{C \sm A}}$ for $A\subseteq C$,  
which to every real-valued function $\phi_C: \Omega_C \ra \sR$
associates the function $\sum_{x_{C \sm A}}\ \phi_C: \Omega_A \ra
\sR$, defined with

\begin{equation}
\Big( \sum_{x_{C\sm A}}\!\!\phi_C\Big) (x_A)=%
\sum_{x_{C \sm A} \in \Omega_{x_{C \sm A}}}\ \phi_C(x_{C})%
\end{equation}
and the marginalization is defined with
\begin{equation}
\phi_C^{\da{A}}=\sum_{x_{C\sm A}}\ \phi_C.
\end{equation}

\subsection{Definition of first order moments}

The joint probability non-negative function of random variable
$X_U$, $p: \Omega_U \ra \sR$ is said to \emph{factorize
multiplicatively} on $\sT$ if there exists non-negative real
functions $p_C: \Omega_C \ra \sR$ for $C \in \sC$, so that we can
write $p(x_U)$ as

\begin{equation}
\label{problem: pU=prod pC}
p_U = %
\prod_{C \in \sC}p_C,
\end{equation}

Similarly, the function $h: \Omega_U \ra \sR$ is said to
\emph{factorize additively} on $\sT$ if there exists real
functions $h_C: \Omega_C \ra \sR$ for $(C \in \sC)$, so that we
can write $h(x_U)$ as

\begin{equation}
\label{problem: hU=sum hC} h_U = \sum_{C \in \sC}h_C
\end{equation}

\emph{The first order moment potential}, $m_C: \Omega_C \ra \sR$,
is defined with
\begin{equation}
m_C = \sum_{x_{U \setminus C}} p_U \cdot h_U.
\end{equation}
In the case $C = \es$, the first order moment potential is simply
denoted $m$,
\begin{equation}
\label{m} m = \sum_{x_U} p_U \cdot h_U.
\end{equation}
and called \emph{the first order moment}.

\begin{example}

The first order moment potential may be useful for expressing the
conditional expectation
\begin{equation}
\mb E \big[ h_U(X_{U}) | x_C \big]=%
\sum_{x_{U \setminus C}} p(X_U | x_C)h(X_{U \setminus C}, x_C)
\end{equation}
for $C \in \sC$. After usage of
\begin{equation}
p(X_{U \setminus C}| x_C) = %
\frac{p(X_{U \setminus C}, x_C)}{p(x_C)}=%
\frac{p(X_{U \setminus C}, x_C)}{\sum_{x_{U \setminus C}} p(X_{U
\setminus C},
x_C)}%
\end{equation}
we have
\begin{equation}
\mb E \big[ h_U(X_{U}) | x_C \big]=%
\frac{\sum_{x_{U \setminus C}} p(X_{U \setminus C}, x_C)h(X_{U
\setminus C}, x_C)}{\sum_{x_{U \setminus C}} p(X_{U \setminus C},
x_C)}= \frac{m_C(x_C)}{p_U^{\da C}(x_C)}.
\end{equation}
Consequently, unconditioned expectation equals the first order
moment
\begin{equation}
\mb E \big[ h(X_U) \big]=%
\sum_{x_U} p(x_{U})h(x_U)= m .
\end{equation}

\end{example}

\subsection{The problem of first order moments computation as all vertices
problem}

The computation of (\ref{m}) by enumerating all configurations
would require an exponential number of operations with respect to
the cardinality of $\Omega_U$. However, the computational
complexity can be reduced using the local computation algorithm
which exploits structure of functions given with factorizations
(\ref{problem: pU=prod pC}) and (\ref{problem: hU=sum hC}). In
this case, the marginal values $p_U^{\da C}$ are computed for all
$C \in \sC$ using the local computation over the set of
real-valued functions. After that the moment is computed according
to equality
\begin{equation}
m_C = \sum_{C \in \sC} \sum_{x_C} h_C \ p_U^{\da C},
\end{equation}
which follows from
\begin{equation}
m_C = \sum_{x_U} p_U \ h_U =%
\sum_{x_U} p_U  \sum_{C \in \sC} h_C = %
\sum_{C \in \sC} \sum_{x_U} p_U \ h_C = %
\sum_{C \in \sC} \sum_{x_C} h_C \sum_{x_{U \sm C}} p_U =%
\sum_{C \in \sC} \sum_{x_C} h_C \ p_U^{\da C}.
\end{equation}
This method requires the storing of marginal values $p_U^{\da C}$
for all $C \in \sC$, which unnecessary increases the memory
complexity. Instead, we can use the local computation algorithms
by Lauritzen and Nilsson \cite{Lauritzen_Nillson_01} and Mau{\'a}
et al. \cite{Maua_et_al_11}, which find the moment as the solution
of the normalization problem. In the following section, we
consider these two algorithms.

\section{First order moments computation using order pair
potential algorithms}

\subsection{Order pair potentials}

In our local computation algorithms we represent the quantitative
elements through entities called potentials. Each such potential
has two parts, as detailed below.

\begin{definition} \textbf{(Potential)} A potential on $C \subseteq U$ is a pair
$\pi_{C}$=$(p_{C}$, $h_{C})$ of real-valued functions on $\Omega_C$,
where $p_{C}$ is nonnegative.
\end{definition}

Thus, a potential consists of two parts - $p$-part and $h$-part.
We call a potential $\pi_C$ \textit{vacuous}, if $\pi_C = (1,0)$.
We identify two potentials $\pi_C^{(1)}=(p_C^{(1)}, h_C^{(1)})$ and $\pi_C^{(2)}=
(p_C^{(2)}, h_C^{(2)})$ on $C$ and write $\pi_C^{(1)}=\pi_C^{(2)}$ if
$p_C^{(1)}=p_C^{(2)}$ and $h_C^{(1)}(x_C)=h_C^{(2)}(x_C)$ whenever
\begin{equation}
p_C^{(1)}(x_C)=p_C^{(2)}(x_C)>0,
\end{equation}
i.e., two potentials are considered equal if they have identical
probability parts and their utility parts agree almost surely with
respect to the probability parts.

To represent and evaluate the decision problem in terms of
potentials, we define basic operations of \textit{combination} and
\textit{marginalization}. There are two possible ways to define the
operations.

\begin{enumerate}

\item Lauritzen-Nilsson algorithm \cite{Lauritzen_Nillson_01}
\item Mau{\'a} et al. algorithm \cite{Maua_et_al_11} 

\end{enumerate}

\subsection{Lauritzen-Nilsson algorithm}

\begin{definition}\textbf{(Combination)} The \textit{combination} of two
potentials $\pi_A = (p_A,h_A)$ and $\pi_B =
(p_B, h_B)$ denotes the potential on $A \cup B$ given by
\begin{equation}
\label{LN_combination}
\pi_A\otimes \pi_B = (p_A\cdot p_B,h_A + h_B).
\end{equation}

\end{definition}

\begin{definition}\textbf{(Marginalization)} The \textit{marginalization}
of $\pi_C = (p_C,h_C)$ onto $A\subseteq C \in \sC$ is defined
by
\begin{equation}
\label{LN_marginalization} \pi_C^{\da A} = \Big( \sum_{x_{C\sm
A}} p_C,\frac{\sum_{x_{C \sm A}}p_C \cdot h_C}{\sum_{x_{C \sm
A}} p_C}\Big)
\end{equation}

\end{definition}

Here we have used the convention that 0/0 which will be used
throughout.

As shown in Lauritzen and Nilsson \cite{Lauritzen_Nillson_01}, the
operations of combination and marginalization satisfy the
properties of Shenoy and Shafer axioms \cite{Shenoy_Shafer_90},
and three structured factorizations can be marginalized using the
Shafer-Shenoy algorithm.

If the operations are defined in this way and the potentials are set
to,
\begin{equation}
\label{LN_phi}
\phi_C = (p_C \ ,\ h_C)
\end{equation}
and the factorizations (\ref{problem: pU=prod pC}) and
(\ref{problem: hU=sum hC}) hold, then
\begin{equation}
\pi_U = \bigotimes_{C \in \sC} \pi_C =%
\bigotimes_{C \in \sC} (p_C, h_C) = \big(\prod_{C \in \sC} p_C,
\sum_{C \in \sC} h_C \big) = ( p_U, h_U ).
\end{equation}
Accordingly, we have
\begin{equation}
\pi_U^{\da \es} =%
\Big( \sum_{x_U} p_U ,%
\frac{\sum_{x_U} p_U \ h_U}{\sum_{x_U} p_U} \Big)=%
\big(1 , m \big),
\end{equation}
where we have used probability condition $\sum_{x_U} p_U = 1$.
Hence, the first order moment potential can be computed using the
Shafer-Shenoy local computation algorithm, where the combination
and marginalization are defined with
(\ref{LN_combination})-(\ref{LN_marginalization}). The messages
have the form:
\begin{equation}
\label{LN_message}
\pi_{A \ra B}=( \pi_{A\ra B}^{(p)}, \pi_{A\ra B}^{(h)} )
\end{equation}
where the $p$ and $h$ part are given with
\begin{equation}
\pi_{A\ra B}^{(p)} = \sum_{x_{A \sm B}}\ p_A \prod_{C \in ne(A)
\sm B} \pi_{C \ra A}^{(p)}
\end{equation}
\begin{equation}
\pi_{A \ra B}^{(h)} = \frac{\sum\limits_{x_{A \sm B}} \ p_A
\prod\limits_{C \in ne(A) \sm B} \pi_{C \ra A}^{(p)}
\cdot \Big( h_A + \sum \limits_{C \in ne(A) \sm B} \pi_{C \ra A}^{(h)} \Big)}
{\sum\limits_{x_{A \sm B}} \ p_A \ \prod\limits_{C \in ne(A) \sm B}
\pi_{C \ra A}^{(p)}}
\end{equation}
which follows from equations
(\ref{Message_A_B}), (\ref{LN_combination}), (\ref{LN_marginalization}), (\ref{LN_phi}) and (\ref{LN_message}) .

\begin{example}
Let $\pi_U$ has the chain factorization
\begin{equation}
\pi_U = \bigotimes_{i=1}^n \pi_{V_i},\quad%
V_i \sim V_{i+1} \ \text{for} \ i= 1, \dots, n-1,
\end{equation}
and let $\pi_{i \ra (i+1)}$ stands as shorthand for the message
$\pi_{V_i \ra V_{i+1}}$. According to chain factorization
$ne(V_i)\sm V_{i+1} = \{V_{i-1}\}$, $p$ and $h$ parts of the
message reduce to:
\begin{equation}
\pi_{i \ra (i+1)}^{(p)} = \sum_{x_{V_i \sm V_{i+1}}}\ p_{V_i} \
\pi_{(i-1) \ra i}^{(p)}
\end{equation}
\begin{equation}
\pi_{i \ra (i+1)}^{(h)} =%
\frac{\sum \limits_{x_{V_i \sm V_{i+1}}} \ \ %
p_{V_i} \ \pi_{(i-1) \ra i}^{(p)} \cdot
\big(\ h_{V_i} + \pi_{(i-1) \ra i}^{(h)}\ \big)}
{\sum\limits_{x_{V_i \sm V_{i+1}}}\
p_{V_i} \ \pi_{(i-1) \ra i}^{(p)}}
\end{equation}

\end{example}

\subsection{Mau{\'a} et al. algorithm}

\begin{definition}\textbf{(Combination)} Let $\pi_A = (p_A,h_A)$ and $\pi_B =
(p_B,h_B)$ be two potentials on $A$ and $B$,
respectively. The \emph{combination} $\pi_A \otimes \pi_B$
of $\pi_A$ and $\pi_B$ is the potential on $A \cup B$
given by
\begin{equation}
\label{Mau_combination}
\pi_A \otimes\pi_B = (p_A p_B\ ,\ h_A p_B +
p_A h_B).
\end{equation}

\end{definition}

\begin{definition}\textbf{(Marginalization)} Let $\pi_C = (p_C,h_C)$ be
a potential on $C$, and let $A\subseteq C$. The
\textit{marginalization} $\pi_C^{\downarrow A}$ of $\pi_C$ onto
$A$ is the potential on $A$ given by

\begin{equation}
\label{Mau_marginalization}
\pi_C^{\downarrow A} = \big(\sum_{x_{C \sm A}}p_C, \sum_{x_{C \sm
A}}h_C\big).
\end{equation}

\end{definition}

The following lemma can be proven by induction
\cite{Ilic_et_al_11}.

\begin{lemma}
Let $\mc N \subseteq \sC$ and $\pi_A =( \pi_A^{(z)}, \pi_A^{(h)})$
be order pair potentials for $A \in \mc N$. Then,
\begin{equation}
\label{Mao_lemma}
\bigotimes_{C \in \mc N} \pi_C =%
\bigotimes_{C \in \mc N} (\pi_C^{(p)}, \pi_C^{(h)})=%
\Big( \prod_{A \in \mc N} \pi_A^{(p)}, %
\sum_{A \in \mc N} \pi_A^{(h)} %
\prod_{B \in \mc N \sm A} \pi_B^{(p)}\Big).
\end{equation}
\end{lemma}

If the operations are defined in this way and the potentials are
set to,
\begin{equation}
\label{Mau_phi} \pi_C = (p_C\ ,\ p_C h_C)
\end{equation}
and the factorizations (\ref{problem: pU=prod pC}) and
(\ref{problem: hU=sum hC}) hold, then
\begin{equation}
\pi_U = \bigotimes_{C \in \sC} \pi_C =%
\big(\prod_{A \in \sC} p_A,%
\prod_{A \in \sC} p_A \sum_{B \in \sC} h_B \big) = ( p_U, p_U h_U
).
\end{equation}
Accordingly, we have
\begin{equation}
\pi_U^{\da \es} =%
\Big( \sum_{x_U} p_U ,%
\sum_{x_U} p_U \ h_U \Big)=%
\big(1 , m \big),
\end{equation}
where we have used probability condition $\sum_{x_U} p_U = 1$.
Again, the first order moment potential can be computed using the
Shafer-Shenoy local computation algorithm, where the combination
and marginalization are defined with (\ref{Mau_combination}) and
(\ref{Mau_marginalization}). Like in the Lauritzen-Nilsson
algorithm, the messages have the form:
\begin{equation}
\label{LN_message} \pi_{A \ra B}=( \pi_{A\ra B}^{(p)}, \pi_{A\ra
B}^{(h)} )
\end{equation}
but now, according to \ref{Mao_lemma}, the $p$-part and the
$h$-part of the messages are given with
\begin{equation}
\label{Mau_message_p-part}
\pi_{A\ra B}^{(p)} = \sum_{x_{A\sm B}} \ p_A \ \prod_{C \in ne(A) \sm B}
\pi_{C \ra A}^{(p)}.
\end{equation}
\begin{equation}
\label{Mau_message_h-part}%
\pi_{A\ra B}^{(h)} =%
\sum_{x_{A\sm B}} p_A \cdot \Big(%
\prod_{C \in ne(A) \sm B} \pi_{C \ra A}^{(p)}\cdot h_A \ \ +
\sum_{C \in ne(A) \sm B} \pi_{C\ra A}^{(h)} \prod_{D \in ne(A) \sm
B,C} \pi_{D \ra A}^{(p)} \Big).
\end{equation}
Note that the $p$-parts of the Lauritzen-Nilsson algorithm and the
Mau{\'a} et al. algorithm are the same. For the trees with large
average degree, the $h$-parts of messages are more complex in
Mau{\'a} et al. algorithm, due to repeated multiplications in
products in the equality (\ref{Mau_message_h-part}). However,
Mau{\'a} et al. algorithm is simpler for chains as the following
example shows.

\begin{example}
Let $\pi_U$ has the chain factorization
\begin{equation}
\pi_U = \bigotimes_{i=1}^n \pi_{V_i},\quad%
V_i \sim V_{i+1} \ \text{for} \ i= 1, \dots, n-1,
\end{equation}
and let $\pi_{i \ra (i+1)}$ stands as shorthand for the message
$\pi_{V_i \ra V_{i+1}}$. According to chain factorization
$ne(V_i)\sm V_{i+1} = \{V_{i-1}\}$, $p$ and $h$ parts of the
message reduce to:

\begin{equation}
\pi_{i \ra (i+1)}^{(p)} = \sum_{x_{V_i \sm V_{i+1}}}\ p_{V_i} \
\pi_{(i-1) \ra i}^{(p)},
\end{equation}
\begin{equation}
\pi_{i \ra (i+1)}^{(h)} =%
\sum_{x_{V_i \sm V_{i+1}}} p_{V_i} \cdot %
\big(\pi_{(i-1) \ra i}^{(p)} h_{V_i}+  \pi_{(i-1)\ra i}^{(h)}
\big).
\end{equation}

\end{example}


\section{Conclusion}

We reviewed some existing methods for the computation of first
order moments on junction trees using Shafer-Shenoy algorithm.
First, we consider the problem of first order moments computation
as vertices problem in junction trees. In this way, the problem is
solved using the memory space of an order of the junction tree
edge-set cardinality. After that, we considered two algorithms,
Lauritzen-Nilsson algorithm, and Mau{\'a} et al. algorithm, which
computes the first order moments as the normalization problem in
junction tree, using the memory space of an order of the junction
tree leaf-set cardinality. It is shown, that for trees, the first
of them has simpler formulas in comparison to the second one,
while the second one is simpler for chains.

\bibliographystyle{unsrt}
\bibliography{TR}

\end{document}